\documentclass[sigconf]{acmart}

\usepackage{graphicx}
\usepackage{multirow}
\usepackage{makecell}
\usepackage{url}
\usepackage{footnote}
\makesavenoteenv{figure}
\usepackage{algorithm}
\usepackage{algpseudocode}
\usepackage{algorithmicx}
\usepackage{xcolor}
\usepackage{multirow}
\usepackage{xcolor}
\usepackage[normalem]{ulem} 

\newcommand{\etal}{et al.}

\AtBeginDocument{%
  }

\setcopyright{cc}
\copyrightyear{2026}
\setcctype{by}
\acmYear{2026}

\acmConference[RichMediaGAI '26]{The fourth International Workshop on Rich Media with Generative AI }{November 10--14, 2026}{Rio de Janeiro, Brazil}
\acmDOI{10.1145/3841458.3841542} 
\acmBooktitle{The fourth International Workshop on Rich Media with Generative AI (RichMediaGAI '26), November 10--14, 2026, Rio de Janeiro, Brazil}

\acmISBN{979-8-4007-2947-8/2026/11} 

\acmSubmissionID{5}

\begin{document}

\title{Direct, Parallel, or Sequential? A Comparative Study of Training-Free Multi-Subject Image-to-Video Generation}



\author{Yanliang Qi}
\affiliation{%
  \institution{University of Memphis}
  \city{Memphis}
  \state{TN}
  \country{USA}
}

\author{Kexi Chen}
\authornote{Work done as a research scholar at the University of Memphis.}
\affiliation{%
  \institution{University of Pennsylvania}
  \city{Philadelphia}
  \state{PA}
  \country{USA}
}

\author{Muchao Ye}
\affiliation{%
  \institution{University of Iowa}
  \city{Iowa City}
  \state{IA}
  \country{USA}
}

\author{Haomiao Ni}
\authornote{Corresponding author: hni@memphis.edu.}
\affiliation{%
  \institution{University of Memphis}
  \city{Memphis}
  \state{TN}
  \country{USA}
}
\renewcommand{\shortauthors}{Qi et al.}
\renewcommand{\shorttitle}{A Comparative Study of Training-Free Multi-Subject Image-to-Video Generation}

\begin{abstract}
Text-conditioned image-to-video (I2V) generation has advanced rapidly, yet generating videos with multiple subjects remains challenging. A model must simultaneously preserve the appearance of each subject, assign distinct motions, and maintain coherent spatial and temporal interactions. This paper presents a systematic study of three representative paradigms for training-free multi-subject I2V generation: direct, parallel, and sequential generation. Direct generation applies a pretrained I2V model to the complete reference image and prompt, requiring all subjects and motions to be synthesized jointly. Parallel and sequential generation instead decompose the reference image and prompt into subject-specific visual and textual conditions. Parallel generation synthesizes each subject independently and subsequently composes the resulting videos, reducing the complexity of each generation step at the cost of weaker inter-subject context. Sequential generation first synthesizes a background video and then progressively introduces individual subjects. This preserves accumulated scene context but introduces sensitivity to subject ordering and error propagation. We empirically evaluate the three paradigms across diverse multi-subject scenes, comparing appearance preservation, motion fidelity, temporal consistency, and inter-subject coherence, while also characterizing their distinct failure modes. Our findings reveal the strengths and limitations of each paradigm and offer practical insights for designing controllable multi-subject video generation systems.
\end{abstract}





\maketitle

\section{Introduction}

Text-conditioned image-to-video (I2V) generation has gained significant attention for its ability to synthesize expressive videos from a single static image and a textual motion prompt \cite{hu2022make}. Formally, given an input image $I$ and a prompt $T$, the task is to generate a video ${V}$ that begins with $I$ and reflects the semantics of $T$. Recent diffusion-based models~\cite{ho2020denoising,song2020denoising,song2020score} have achieved impressive single-subject video generation, supporting applications in entertainment, virtual reality, and human-computer interaction \cite{kong2024hunyuanvideo,zhang2023i2vgen}.

\begin{figure}[t]
    \centering
   \includegraphics[width=0.85\linewidth]{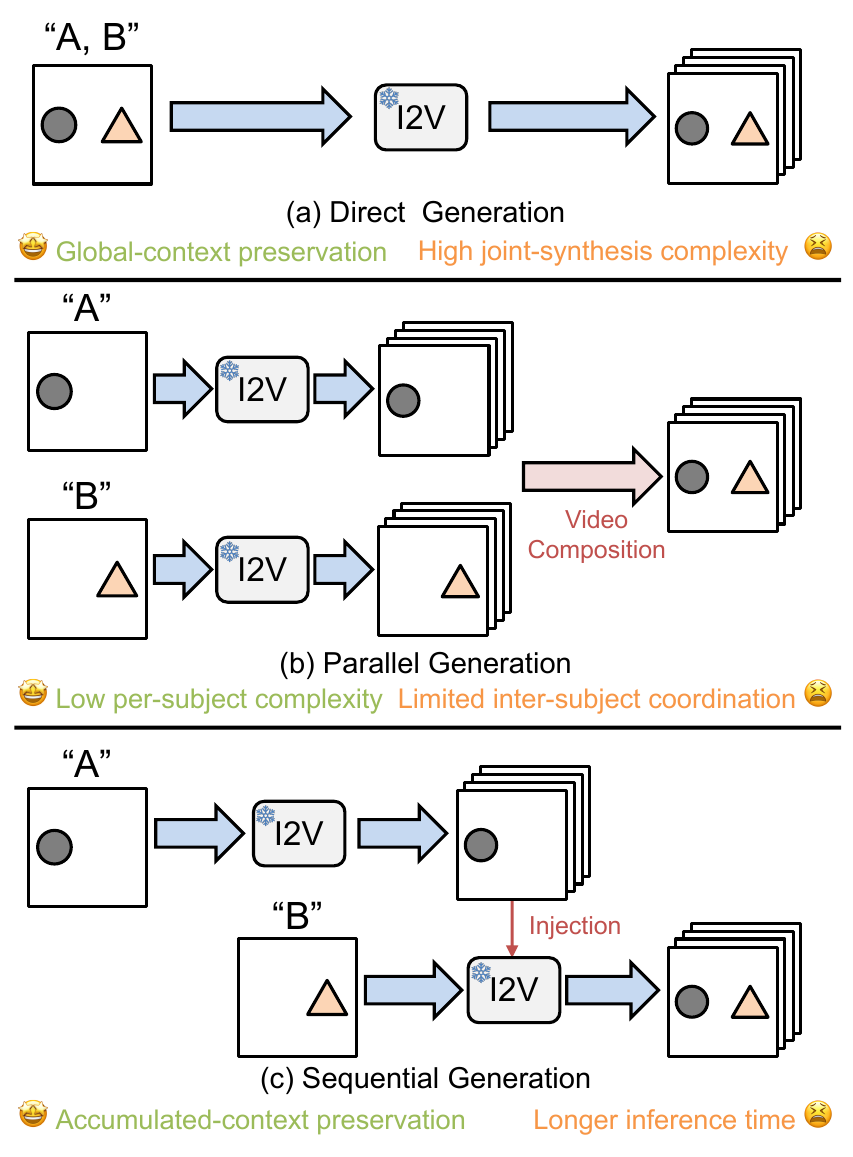}
    \caption{
    Illustration of three training-free strategies for multi-subject image-to-video generation. Direct generation synthesizes all subjects jointly in one pass; parallel generation independently generates subject-level videos and merges them post hoc; sequential generation injects subjects one by one while conditioning on previously generated video context. The pretrained image-to-video (I2V) models remain frozen across all methods. For simplicity, the decomposition of the input multi-subject image and text prompt is omitted in (b) and (c).}
    \label{fig:motivation}
    \Description{Figure described in the caption.}
\end{figure}

Despite these successes, multi-subject I2V generation, in which distinct subjects may follow different prompt-specified motions, remains underexplored. Preserving multiple subject appearances while following individual and contextual motion descriptions remains challenging, raising the question of how frozen pretrained I2V models can support multi-subject generation without additional training. To address this question, we investigate three representative training-free paradigms: direct, parallel, and sequential generation.

As shown in Fig.~\ref{fig:motivation}(a), a direct strategy applies existing I2V models \cite{xing2024dynamicrafter,yang2024cogvideox,zhang2023i2vgen} to generate the entire scene and all subject motions in a single generation process.
This strategy is simple and preserves the complete image–prompt context and overall scene semantics.
However, jointly preserving multiple identities, synthesizing distinct motions, and maintaining realistic inter-subject relationships places a high burden on the model and can produce entangled appearance, motion, or interaction artifacts. 

Parallel generation, illustrated in Fig.~\ref{fig:motivation}(b), alleviates this joint synthesis complexity through decomposition. It decomposes the image and prompt into subject-specific pairs, generates their videos in \textit{parallel} with a pretrained I2V model, and merges the resulting videos using video composition techniques. This strategy reduces the per-subject generation burden and can better preserve individual appearance and motion. However, inter-subject context is not jointly modeled during the independent generation stage. The post-hoc composition stage must therefore reconcile independently generated motions, spatial layouts, occlusions, and interactions, which may produce inconsistent or physically implausible videos.

The third strategy is sequential generation, which introduces subjects progressively while propagating the previously generated context. As illustrated in Fig.~\ref{fig:motivation}(c), each stage introduces one new subject while conditioning on the video containing previously synthesized subjects, reducing the synthesis burden while preserving evolving context. However, its progressive generation increases inference time and may propagate ordering-dependent or accumulated errors.

We compare direct, parallel, and sequential training-free multi-subject I2V generation on a self-created dataset under a unified experimental protocol. Our results show that no paradigm is universally superior: direct generation is simple and efficient but faces high joint synthesis complexity; parallel generation reduces per-subject complexity but loses inter-subject context; and sequential generation preserves progressively accumulated context but increases inference time and is susceptible to error propagation.

\section{Related Work}
\textbf{Image-to-Video Generation.}
Based on the availability of motion cues, image-to-video (I2V) generation can be categorized into \emph{stochastic methods}, which rely solely on a given image $I$ \cite{babaeizadeh2017stochastic}, and \emph{conditional methods}, which utilize both the image $I$ and an external condition $T$ \cite{ni2023conditional}. In this work, we primarily focus on text-conditioned I2V generation \cite{hacohen2024ltx,ni2024ti2v,zhang2023i2vgen}. Hu \etal~\cite{hu2022make} introduced MAGE, a text-driven I2V generator that employs a motion anchor structure to capture appearance-motion-aligned representations using three-dimensional axial transformers. Zhang \etal~\cite{zhang2023i2vgen} proposed I2VGen-XL, a cascaded framework consisting of two stages: a base stage that preserves input semantics through dual hierarchical encoders, and a refinement stage that incorporates the text prompt to enhance visual details and improve video resolution. While these general-purpose I2V frameworks can produce promising results, they are not specifically designed for multi-subject scenarios and often struggle in such settings due to the need to simultaneously synthesize the appearance and motion of multiple subjects. This limitation motivates our study of how pretrained I2V models can be used under different training-free multi-subject generation strategies.

\noindent\textbf{Multi-Subject Video Generation.}
Recent work on multi-subject video generation primarily focuses on customized multi-subject text-to-video (T2V) generation \cite{chen2023videodreamer,chen2024disenstudio,chen2025multi,deng2025cinema,wang2024customvideo}. Chen \etal~\cite{chen2023videodreamer} introduced VideoDreamer, which utilizes a combination of Disen-Mix fine-tuning and human-in-the-loop re-fine-tuning to adapt a pretrained T2V generator for multiple subjects. Chen \etal~\cite{chen2025multi} proposed Video Alchemist, a diffusion-transformer-based framework that fuses each conditional reference image with its corresponding subject-level text prompt via cross-attention layers. 
Compared to multi-subject T2V, multi-subject I2V is more challenging due to the additional constraints imposed by the input image. The generation must not only synthesize coherent motions for multiple subjects but also preserve their appearances, spatial positioning, and inter-subject relationships as defined by the provided image. In contrast to these customization-oriented T2V methods, our work focuses on training-free multi-subject I2V and compares direct, parallel, and sequential generation strategies using frozen pretrained components.

\begin{figure*}[t]
    \centering
    \includegraphics[width=0.85\linewidth]{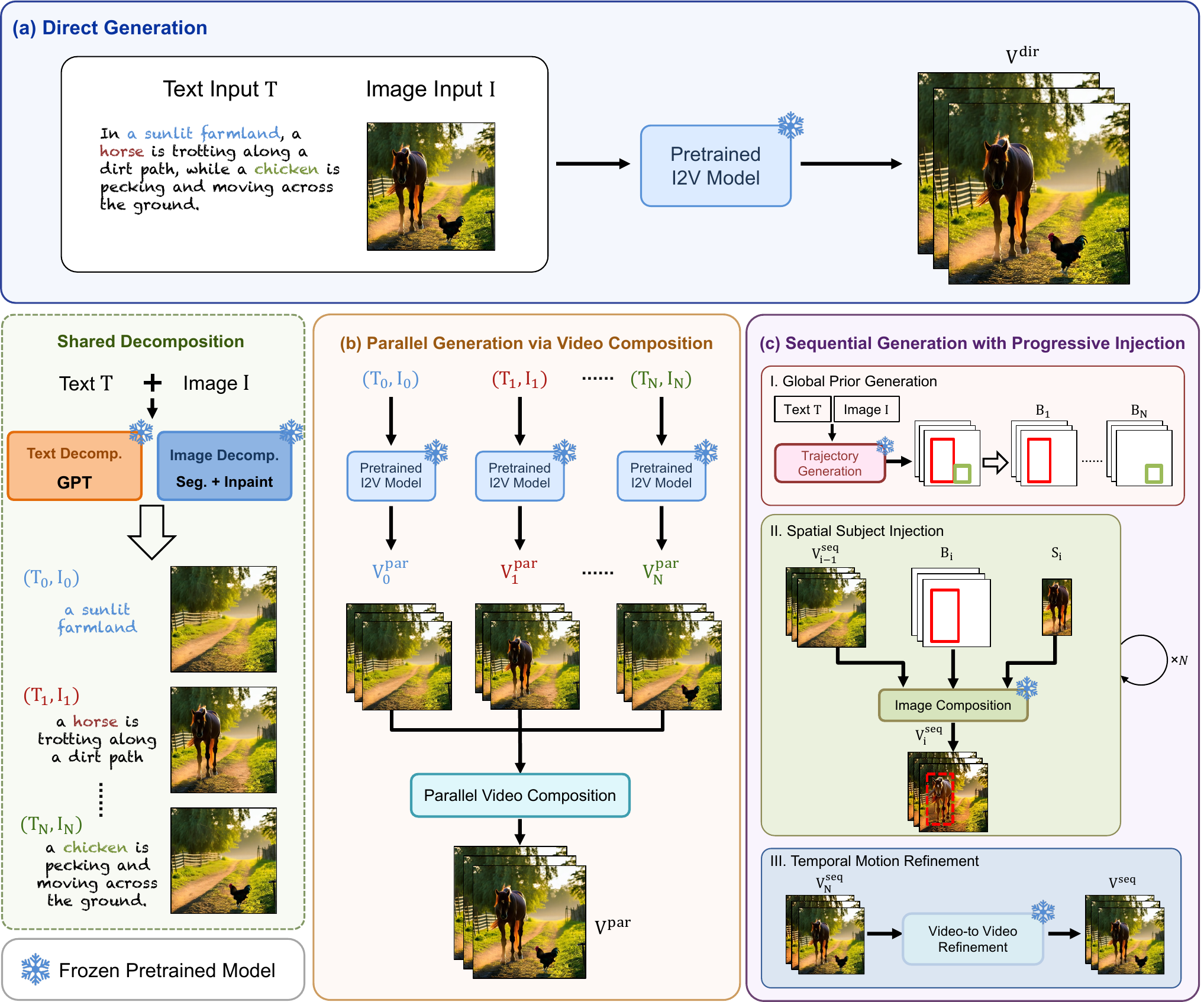}
    \caption{
    Overview of the compared training-free paradigms for multi-subject image-to-video generation. We compare three strategies: (a) \emph{direct generation}, which uses the original image and prompt in a single call to a frozen I2V model; (b) \emph{parallel generation via video composition}, which independently generates decomposed background and subject videos and subsequently composes them; and (c) \emph{sequential generation with progressive injection}, which inserts subjects one by one into an evolving video. For the decomposed paradigms in (b) and (c), the input prompt $T$ and image $I$ are decomposed into a background pair $(T_0,I_0)$ and subject-specific pairs ${(T_i,I_i)}_{i=1}^N$. In (c), stages I, II, and III denote global prior generation, spatial subject injection, and temporal motion refinement, respectively.
    }
    \Description{Figure described in the caption.}
    \label{fig:methodology}
\end{figure*}

\section{Methodology}
\subsection{Problem Formulation and Paradigm Overview}
\label{sec:problem_formulation}
We study training-free multi-subject image-to-video (I2V) generation as a comparison of three generation paradigms: direct generation, parallel generation, and sequential generation.
Formally, given an input image $I$ containing $N$ subjects $\{S_i\}_{i=1}^{N}$ and a text prompt $T$ describing their motions and scene context, the goal is to synthesize a video sequence $V=\langle v^1,\ldots,v^L\rangle$ with $L$ frames.
The generated video should preserve the appearance of each subject, follow the motion described in the prompt, and maintain coherent spatial and temporal relationships among subjects and the background.

As shown in Fig.~\ref{fig:methodology}, the three paradigms organize the generation process in different ways.
Direct generation applies the pretrained I2V model $\mathcal{M}$ to the full image and prompt in one pass, requiring the model to jointly synthesize all subjects and their motions.
Parallel generation first decomposes the input image and prompt into background-level and subject-level components, generates each component independently, and then composes the resulting videos into a multi-subject video.
Sequential generation also uses decomposed components, but injects subjects one by one into a shared evolving video context, followed by temporal refinement.
Below, we first introduce direct generation, then describe the image and text-prompt decomposition, and finally present the parallel and sequential generation paradigms built upon this decomposition.

\subsection{Direct Generation} 
\label{sec:direct_generation} 
Direct generation is the most straightforward strategy for multi-subject I2V generation. It directly applies the pretrained I2V model $\mathcal{M}$ to the full text prompt $T$ and the original multi-subject image $I$: 
\begin{small}
\begin{equation} 
\label{eq:dir}
V^{\mathrm{dir}} = \mathcal{M}(T,I). 
\end{equation} 
\end{small}%
Unlike parallel and sequential generation, this paradigm does not perform text-image decomposition or post-hoc composition. Thus, the model can access the complete image-prompt context throughout generation, preserving the original spatial layout and global semantic description. 

However, direct generation requires $\mathcal{M}$ to synthesize all subjects, their individual motions, and their inter-subject relationships in a single pass. This joint synthesis process places a high burden on the frozen I2V model and may lead to identity confusion, motion entanglement, missing subjects, or weak inter-subject interactions. Therefore, direct generation serves as a simple but challenging reference paradigm for evaluating multi-subject I2V generation.

\subsection{Image and Text Prompt Decomposition}
\label{sec:decomp}
Direct generation operates on the original input image and prompt, while parallel and sequential generation both rely on decomposed text-image components.
Inspired by the ``divide-and-conquer'' principle, we first decompose both the text prompt $T$ and the input image $I$ into modular components that can later be recombined in a controlled manner. 

This shared decomposition step provides a common input representation for the two decomposed paradigms, ensuring that their differences mainly come from the subsequent generation process rather than from different preprocessing procedures.
In particular, text decomposition is performed before image decomposition, since subject identities are first established from the prompt and then used to guide the image decomposition module to isolate each subject.

\noindent\textbf{Text Decomposition.} Given the input text prompt $T$ as global guidance, we decompose it into subject-level descriptions $\{T_i\}_{i=1}^{N}$ and a scene-level description $T_0$ via a large language model (LLM) $\mathcal{W}$: 
\begin{small}
\begin{equation} 
\{T_i\}_{i=0}^{N} = \mathcal{W}(T). 
\end{equation}
\end{small}%
The subject-level prompt $T_i$ specifies the identity, attributes, and motion of subject $S_i$, while removing descriptions of the other subjects as much as possible. The scene-level prompt $T_0$ captures the background environment and global scene context.
This decomposition establishes an explicit correspondence between textual subjects and visual subjects before the image is processed.

\noindent\textbf{Image Decomposition.}  
With $\{T_i\}_{i=0}^{N}$ obtained, we apply an image decomposition pipeline $\mathcal{P}$ to the input image $I$: 
\begin{small}
\begin{equation} 
\{I_i\}_{i=0}^{N} = \mathcal{P}(I,\{T_i\}_{i=1}^{N}). 
\end{equation}
\end{small}%

Specifically, we first apply pretrained object detection and segmentation models to identify each subject in the input image $I$, yielding subject masks and corresponding bounding boxes. 
To obtain the background image $I_0$, all foreground subjects are removed and the missing regions are reconstructed using a pretrained inpainting model.
To obtain single-subject images, we mask out all other subjects through segmentation masks and use a pretrained inpainting model to reconstruct the background in the masked regions. Finally, we crop the preserved subject region using the bounding box, producing a set of single-subject images $\{I_i\}^N_{i=1}$ and a clean background image $I_0$ (with all subjects removed). The resulting text-image pairs $\{(T_i,I_i)\}_{i=0}^{N}$ are then used as the shared decomposed inputs for parallel and sequential generation.

\subsection{Parallel Generation via Video Composition}
\label{sec:parallel_generation}
Parallel generation first synthesizes decomposed components independently and then composes them into a multi-subject video.
Given the decomposed text-image pairs $\{(T_i,I_i)\}_{i=0}^{N}$, we independently generate a background video and $N$ subject-level videos:
\begin{small}
\begin{equation}
    V_i^{\mathrm{par}}=\mathcal{M}(T_i,I_i),\quad i=0,1,\ldots,N .
\end{equation}
\end{small}%
Here, $V_0^{\mathrm{par}}$ denotes the generated background video, and $V_i^{\mathrm{par}}$ denotes the generated video object for subject $S_i$.
Thus, each I2V call only handles one subject or the background.

To merge the independently generated component videos, we adopt a training-free video object composition module from~\cite{wang2024mvoc}.
Different from the original video object composition setting, where the inputs are multiple source video objects, our parallel paradigm first converts each decomposed text-image pair $(T_i,I_i)$ into a generated component video $V_i^{\mathrm{par}}$. The resulting component videos are then provided to the composition module as the source videos corresponding to the individual subjects.

Following~\cite{wang2024mvoc}, for each generated component video $V_i^{\mathrm{par}}$, we first use the segmentation model $\mathcal{Q}$ to extract a sequence of masks for the corresponding subject:
\begin{small}
\begin{equation}
    M_i^{\mathrm{par}} = \mathcal{Q}(V_i^{\mathrm{par}}).
\end{equation}
\end{small}%
We also apply DDIM inversion~\cite{song2020denoising} to obtain its inverted latent trajectory:
\begin{small}
\begin{equation}
    Z_i^{\mathrm{par}} = \operatorname{DDIMInv}(V_i^{\mathrm{par}}).
\end{equation}
\end{small}%
The masks localize object regions, while the inverted latents preserve motion and appearance cues.

The composition module $\mathcal{C}$ then utilizes the original image and prompt, the object masks, and the inverted latents to generate the final multi-subject video:
\begin{small}
\begin{equation}
    V^{\mathrm{par}}
    = \mathcal{C}(T, I,\{Z_i^{\mathrm{par}},M_i^{\mathrm{par}}\}_{i=0}^{N}).
\end{equation}
\end{small}%
During composition, $\mathcal{C}$ leverages feature and attention guidance from the component videos to preserve subject identity and motion consistency in the final video. Further details can be found in~\cite{wang2024mvoc}.

\begin{small}
\begin{algorithm}[t]
    \caption{Parallel generation via video composition.}
    \label{alg:parallel}
    \begin{algorithmic}[1]
    \algrenewcommand\algorithmiccomment[1]{\hfill\textcolor{gray}{// #1}}

    \Require Decomposed text-image pairs $\{(T_i,I_i)\}_{i=0}^{N}$;
    global prompt $T$;
    pretrained I2V model $\mathcal{M}$;
    mask extraction module $\mathcal{Q}$;
    video composition module $\mathcal{C}$.
    \Ensure A synthesized video $V^{\mathrm{par}}$ with $L$ frames.

    \For{$i=0,1,\cdots,N$}
        \State $V_i^{\mathrm{par}} \gets \mathcal{M}(T_i,I_i)$
        \Comment{Independent I2V generation}
        \State $M_i^{\mathrm{par}} \gets \mathcal{Q}(V_i^{\mathrm{par}})$
        \Comment{Mask extraction}
        \State $Z_i^{\mathrm{par}} \gets \operatorname{DDIMInv}(V_i^{\mathrm{par}})$
        \Comment{Latent inversion}
    \EndFor

    \State $V^{\mathrm{par}} \gets
    \mathcal{C}(T,I,\{Z_i^{\mathrm{par}},M_i^{\mathrm{par}}\}_{i=0}^{N})$
    \Comment{Video object composition}

    \State \Return $V^{\mathrm{par}}$
    \end{algorithmic}
\end{algorithm}
\end{small}%
The advantage of parallel generation lies in its modularity: each I2V call handles only one subject or the background, simplifying appearance preservation and motion synthesis. However, independently generated subject videos do not model inter-subject context during generation.
Consequently, spatial compatibility, occlusion, shared shadows, and coordinated interactions are difficult to reconstruct during post-hoc composition, making this paradigm less reliable for strongly interacting subjects.

\subsection{Sequential Generation with Progressive Injection}  
Sequential generation progressively inserts decomposed subjects into a shared, evolving video state, thereby retaining more visual context than independent parallel generation while avoiding the joint synthesis of all subjects in a single pass. Inspired by coarse-to-fine video generation and object insertion frameworks~\cite{ho2022imagen,zhang2023i2vgen,zhao2025dreaminsert}, we instantiate this paradigm as a cascaded pipeline comprising global prior generation, spatial subject injection, and temporal motion refinement. All stages use off-the-shelf pretrained models without additional training or fine-tuning.

\noindent\textbf{Global prior generation.}
We first obtain a global motion prior from the same frozen I2V backbone used in direct generation.
Given the original multi-subject image $I$ and the full prompt $T$, the backbone generates a direct-generation video $V^{\mathrm{dir}}$ using Eq.~\ref{eq:dir}.
Based on $V^{\mathrm{dir}}$, the global prior generation module $\mathcal{G}$ produces an initial background video and subject-level trajectories:
\begin{small}
\begin{equation}
    V_0^{\mathrm{seq}},\{\mathcal{B}_i\}_{i=1}^{N}
    =
    \mathcal{G}(V^{\mathrm{dir}}).
\end{equation}
\end{small}%
Here, $V_0^{\mathrm{seq}}=\langle v_{0}^1,v_{0}^{2},\ldots,v_{0}^{L}\rangle$ denotes the background video obtained by removing foreground subjects from the direct-generation video $V^{\mathrm{dir}}$, and $\mathcal{B}_i=\langle b_i^1,b_i^2,\ldots,b_i^L\rangle $ denotes the bounding-box sequence of subject $S_i$.
These trajectories are used as coarse spatiotemporal anchors for later insertion.

\noindent\textbf{Spatial subject injection.}
Starting from $V_0^{\mathrm{seq}}$, subjects are inserted sequentially according to a predefined order.
At step $i$, the current video state $V_{i-1}^{\mathrm{seq}}$ already contains subjects $\{S_j\}_{j=1}^{i-1}$. 
Given the image $I_i$ of the new subject $S_i$, its trajectory $\mathcal{B}_i$, and the $t$-th frame $v_{i-1}^{t}$ of $V_{i-1}^{\mathrm{seq}}$, a pretrained image composition model $\mathcal{H}$ produces the updated frame containing $S_i$:
\begin{small}
\begin{equation}
    v_{i}^{t}
    =
    \mathcal{H}(I_i,b_i^t,v_{i-1}^{t}),
\end{equation}
\end{small}%
where $b_i^t$ specifies the spatial location of subject $S_i$ in frame $t$.
Applying this operation across all frames gives
\begin{small}
\begin{equation}
    V_i^{\mathrm{seq}}
    =
    \langle v_{i}^1,v_{i}^2,\ldots,v_{i}^L\rangle .
\end{equation}
\end{small}%
After all $N$ subjects are inserted, the procedure produces a coarse multi-subject video $V_N^{\mathrm{seq}}$.
This stage explicitly controls subject placement and preserves the previously generated visual context, but its frame-wise nature may introduce temporal artifacts.

\begin{small}
\begin{algorithm}[t]
    \caption{Sequential generation with progressive injection.}
    \label{alg:sequential}
    \begin{algorithmic}[1]
    \algrenewcommand\algorithmiccomment[1]{\hfill\textcolor{gray}{// #1}}

    \Require Original image $I$ with $N$ subjects;
    global prompt $T$;
    decomposed images $\{I_i\}_{i=0}^{N}$;
    pretrained I2V model $\mathcal{M}$;
    global prior module $\mathcal{G}$;
    image composition model $\mathcal{H}$;
    temporal refinement model $\mathcal{R}$.
    \Ensure A synthesized video $V^{\mathrm{seq}}$ with $L$ frames.

    \State $V^{\mathrm{dir}} \gets \mathcal{M}(T,I)$
    \Comment{Direct-generation motion prior}
    \State $V_0^{\mathrm{seq}},\{\mathcal{B}_i\}_{i=1}^{N}
    \gets \mathcal{G}(V^{\mathrm{dir}})$
    \Comment{Background video and trajectories}

    \For{$i=1,2,\cdots,N$}
        \For{$t=1,2,\cdots,L$}
            \State $v_i^t
            \gets \mathcal{H}(I_i,b_i^t,v_{i-1}^t)$
            \Comment{Spatial composition}
        \EndFor
    \EndFor

    \State $V^{\mathrm{seq}} \gets \mathcal{R}(V_N^{\mathrm{seq}},T,I)$
    \Comment{Temporal refinement}
    \State \Return $V^{\mathrm{seq}}$
    \end{algorithmic}
\end{algorithm}
\end{small}%

\noindent\textbf{Temporal motion refinement.}
Since the coarse video $V_N^{\mathrm{seq}}$ is obtained through frame-level composition, it may contain inter-frame inconsistency, boundary artifacts, or unnatural local motion.
We therefore apply a pretrained text- and image-conditioned video-to-video (V2V) refinement model $\mathcal{R}$:
\begin{small}
\begin{equation}
    V^{\mathrm{seq}}
    = \mathcal{R}\!\left(V_N^{\mathrm{seq}}, T,I\right).
\end{equation}
\end{small}%
Here, $I$ and $T$ denote the original multi-subject image and global text prompt, respectively. This refinement stage uses the temporal prior of the pretrained V2V model to improve motion smoothness and visual continuity while preserving the composed subject layout.

Overall, sequential generation represents a middle ground between direct and parallel paradigms.
By progressively updating the intermediate video state $V_i^{\mathrm{seq}}$, it reduces the burden of synthesizing all subjects simultaneously while retaining more shared visual context than fully independent parallel generation.
This cumulative context allows later inserted subjects to be composed with previously inserted ones, which can better preserve spatial compatibility and local interactions.
However, sequential generation is inherently order-dependent: subjects inserted earlier are generated without awareness of those introduced later, which may limit coordinated motion and interaction modeling. In addition, errors from early insertion steps may propagate and accumulate throughout the sequence. The final output is therefore sensitive to the insertion order, the quality of the extracted trajectories, frame-level composition errors, and the robustness of the temporal refinement module.

\section{Experiments}
We conduct extensive experiments to compare direct, parallel, and sequential generation for multi-subject image-to-video synthesis. We describe the self-constructed benchmark, evaluation metrics, and implementation protocol, then analyze the three paradigms through quantitative results, qualitative comparisons, ablation studies, efficiency evaluation, and failure cases.

\subsection{Dataset and Metrics}
To comprehensively compare different paradigms for multi-subject I2V generation, we construct a controlled multi-subject dataset. The construction process involves two steps: (1) \emph{Text Prompt Generation.} We manually design text prompts that describe natural scenes involving two subjects with explicit motion descriptions. This choice emphasizes the multi-subject nature of our task while keeping the scenarios interpretable and controlled. (2) \emph{Image Prompt Generation.} Given each text prompt, we use Stable Diffusion 3.5-Large~\cite{rombach2022high} to synthesize a corresponding multi-subject image at the resolution of $512\times512$. Low-quality outputs are manually filtered out, resulting in a dataset of 20 high-quality text--image pairs. All three paradigms, including direct, parallel, and sequential generation, are evaluated on the same dataset to ensure a fair comparison.

For evaluation, we adopt five representative metrics from VBench \cite{huang2024vbench}, including: {subject consistency} (SC), {background consistency} (BC), {motion smoothness} (MS), {aesthetic quality} (AQ), and {imaging quality} (IQ).

\subsection{Implementation Details}
\noindent\textbf{Compared Paradigms and Backbones.}
We implement and compare three paradigms for multi-subject I2V generation: direct, parallel, and sequential generation. To examine whether the comparison is consistent across different pretrained I2V models, we evaluate the three paradigms with two backbones, I2VGen-XL~\cite{zhang2023i2vgen} and DynamiCrafter~\cite{xing2024dynamicrafter}. Unless otherwise specified, I2VGen-XL is used as the default I2V backbone $\mathcal{M}$ due to its strong video synthesis capability, and the guidance scale $g$ is set to 9.

\begin{table*}[t]
\centering
\resizebox{0.55\textwidth}{!}{%
\begin{tabular}{cccc|ccccc}
\hline
\multicolumn{4}{c|}{\textbf{Setting}}                                              & \multicolumn{5}{c}{\textbf{Metrics}}                                                    \\ \hline
\multicolumn{1}{c|}{\textbf{Paradigm}}           & \multicolumn{1}{c|}{\textbf{BB}}          & \multicolumn{1}{c|}{\textbf{CFG}}       & \textbf{Steps} & \textbf{SC(\%)} & \textbf{BC(\%)} & \textbf{MS(\%)} & \textbf{AQ(\%)} & \textbf{IQ(\%)} \\ \hline
\multicolumn{1}{c|}{\multirow{8}{*}{Direct}}     & \multicolumn{1}{c|}{\multirow{4}{*}{DC}}  & \multicolumn{1}{c|}{\multirow{2}{*}{7}} & 25             & \textbf{92.34}  & 95.15           & 97.38           & 60.11           & 57.25           \\
\multicolumn{1}{c|}{}                            & \multicolumn{1}{c|}{}                     & \multicolumn{1}{c|}{}                   & 50             & 91.54           & \textbf{95.20}  & 96.87           & 60.56           & 57.83           \\ \cline{3-4} 
\multicolumn{1}{c|}{}                            & \multicolumn{1}{c|}{}                     & \multicolumn{1}{c|}{\multirow{2}{*}{9}} & 25             & 91.36           & 94.45           & 97.16           & 58.80           & 56.90           \\
\multicolumn{1}{c|}{}                            & \multicolumn{1}{c|}{}                     & \multicolumn{1}{c|}{}                   & 50             & 90.91           & 94.78           & 96.81           & 59.43           & 56.53           \\ \cline{2-9} 
\multicolumn{1}{c|}{}                            & \multicolumn{1}{c|}{\multirow{4}{*}{I2V}} & \multicolumn{1}{c|}{\multirow{2}{*}{7}} & 25             & 85.37           & 90.27           & 95.83           & 58.07           & 71.26           \\
\multicolumn{1}{c|}{}                            & \multicolumn{1}{c|}{}                     & \multicolumn{1}{c|}{}                   & 50             & 91.43           & 94.66           & \textbf{98.49}  & \textbf{64.28}  & \textbf{73.48}  \\ \cline{3-4} 
\multicolumn{1}{c|}{}                            & \multicolumn{1}{c|}{}                     & \multicolumn{1}{c|}{\multirow{2}{*}{9}} & 25             & 85.34           & 90.63           & 95.97           & 58.90           & 70.26           \\
\multicolumn{1}{c|}{}                            & \multicolumn{1}{c|}{}                     & \multicolumn{1}{c|}{}                   & 50             & 91.26           & 94.79           & 98.44           & 64.05           & 72.51           \\ \hline
\multicolumn{1}{c|}{\multirow{8}{*}{Parallel}}   & \multicolumn{1}{c|}{\multirow{4}{*}{DC}}  & \multicolumn{1}{c|}{\multirow{2}{*}{7}} & 25             & 89.56           & 93.72           & 96.14           & 58.06           & 66.62           \\
\multicolumn{1}{c|}{}                            & \multicolumn{1}{c|}{}                     & \multicolumn{1}{c|}{}                   & 50             & 87.64           & 92.83           & 94.74           & 59.81           & 68.04           \\ \cline{3-4} 
\multicolumn{1}{c|}{}                            & \multicolumn{1}{c|}{}                     & \multicolumn{1}{c|}{\multirow{2}{*}{9}} & 25             & 89.75           & 93.59           & 95.39           & 58.82           & 68.09           \\
\multicolumn{1}{c|}{}                            & \multicolumn{1}{c|}{}                     & \multicolumn{1}{c|}{}                   & 50             & 87.35           & 92.64           & 94.31           & 58.05           & 68.35           \\ \cline{2-9} 
\multicolumn{1}{c|}{}                            & \multicolumn{1}{c|}{\multirow{4}{*}{I2V}} & \multicolumn{1}{c|}{\multirow{2}{*}{7}} & 25             & 85.50           & 90.97           & 94.35           & 56.67           & \textbf{71.86}  \\
\multicolumn{1}{c|}{}                            & \multicolumn{1}{c|}{}                     & \multicolumn{1}{c|}{}                   & 50             & \textbf{91.77}  & \textbf{94.17}  & \textbf{96.34}  & \textbf{61.05}  & 71.69           \\ \cline{3-4} 
\multicolumn{1}{c|}{}                            & \multicolumn{1}{c|}{}                     & \multicolumn{1}{c|}{\multirow{2}{*}{9}} & 25             & 85.57           & 90.73           & 94.29           & 56.44           & 69.91           \\
\multicolumn{1}{c|}{}                            & \multicolumn{1}{c|}{}                     & \multicolumn{1}{c|}{}                   & 50             & 90.32           & 93.58           & 96.18           & 59.76           & 70.03           \\ \hline
\multicolumn{1}{c|}{\multirow{8}{*}{Sequential}} & \multicolumn{1}{c|}{\multirow{4}{*}{DC}}  & \multicolumn{1}{c|}{\multirow{2}{*}{7}} & 25             & 93.41           & \textbf{95.49}  & 96.73           & \textbf{65.88}  & \textbf{73.93}  \\
\multicolumn{1}{c|}{}                            & \multicolumn{1}{c|}{}                     & \multicolumn{1}{c|}{}                   & 50             & \textbf{94.13}  & 95.46           & 96.71           & 64.15           & 73.77           \\ \cline{3-4} 
\multicolumn{1}{c|}{}                            & \multicolumn{1}{c|}{}                     & \multicolumn{1}{c|}{\multirow{2}{*}{9}} & 25             & 93.25           & 94.83           & 95.87           & 63.03           & 72.28           \\
\multicolumn{1}{c|}{}                            & \multicolumn{1}{c|}{}                     & \multicolumn{1}{c|}{}                   & 50             & 92.69           & 95.00           & \textbf{96.75}  & 61.73           & 72.13           \\ \cline{2-9} 
\multicolumn{1}{c|}{}                            & \multicolumn{1}{c|}{\multirow{4}{*}{I2V}} & \multicolumn{1}{c|}{\multirow{2}{*}{7}} & 25             & 91.50           & 93.74           & 94.59           & 59.83           & 71.64           \\
\multicolumn{1}{c|}{}                            & \multicolumn{1}{c|}{}                     & \multicolumn{1}{c|}{}                   & 50             & 93.46           & 95.03           & 95.74           & 60.01           & 70.27           \\ \cline{3-4} 
\multicolumn{1}{c|}{}                            & \multicolumn{1}{c|}{}                     & \multicolumn{1}{c|}{\multirow{2}{*}{9}} & 25             & 91.43           & 93.83           & 93.91           & 58.53           & 68.06           \\
\multicolumn{1}{c|}{}                            & \multicolumn{1}{c|}{}                     & \multicolumn{1}{c|}{}                   & 50             & 93.98           & 94.94           & 95.69           & 59.77           & 69.31           \\ \hline
\end{tabular}
}
\caption{
Quantitative comparison of three multi-subject I2V generation paradigms under different backbone and sampling settings.
\textbf{BB} denotes the pretrained I2V backbone, where \textbf{DC} and \textbf{I2V} refer to DynamiCrafter and I2VGen-XL, respectively.
\textbf{CFG} denotes the classifier-free guidance scale $g$, and \textbf{Steps} denotes the number of denoising steps used in the I2V generation stage.
We report five VBench-style metrics: subject consistency (SC), background consistency (BC), motion smoothness (MS), aesthetic quality (AQ), and imaging quality (IQ).
All scores are reported as percentages, where higher values indicate better performance.
Within each paradigm, the best score over different backbones, CFG, and step settings is highlighted in bold.
}
\label{tab:compare_three}
\end{table*}

\noindent\textbf{Shared Decomposition Modules.}
We use the same text and image decomposition pipeline introduced in Sec.~\ref{sec:decomp} for both Parallel Generation and Sequential Generation. For text decomposition, we employ GPT-4~\cite{achiam2023gpt} as $\mathcal{W}$ to decompose the original prompt $T$ into a set of prompts $\{T_i\}_{i=0}^{N}$, where $T_0$ describes the background and $T_i$ describes subject $S_i$ for $i=1,\ldots,N$. For image decomposition, we use Grounding DINO~\cite{liu2024grounding} for object detection and SAM~\cite{kirillov2023segment} for segmentation, while background inpainting is carried out by Stable Diffusion-1.5~\cite{rombach2022high}. This shared decomposition design ensures that the comparison between parallel and sequential generation is not affected by different decomposition modules.

\noindent\textbf{Parallel Generation.}
Given the decomposed text-image pairs, background and subject videos are independently generated using the backbone, guidance scale, and denoising steps specified in Table~\ref{tab:compare_three}. 
The mask extraction module $\mathcal{Q}$ is instantiated as a fully pretrained Grounded-SAM-2 pipeline~\cite{ren2024grounded}. Specifically, for each independently generated subject video, Grounding DINO Tiny~\cite{liu2024grounding} detects candidate object boxes in the first frame using the corresponding subject prompt, with box and text thresholds set to $0.20$ and $0.15$, respectively. CLIP ViT-L/14~\cite{radford2021learning} re-ranks the candidate boxes according to text--image similarity, and the selected box is provided as the initial prompt to SAM 2.1 Hiera Large~\cite{ravi2025sam}, which extracts the first-frame instance mask and propagates it through the remaining video frames to produce masks for the whole video. No manually annotated masks or additional training are used.
Each video branch is subsequently inverted into DDIM latents using 500 inversion steps with $g=1.0$. We then employ MVOC~\cite{wang2024mvoc} as the video composition module $\mathcal{C}$ to compose the inverted branches through mask-guided Plug-and-Play feature and attention injection~\cite{tumanyan2023plug}, using 50 denoising steps with $g=9.0$. All remaining composition hyperparameters follow the official MVOC defaults.

\noindent\textbf{Sequential Generation.}
The image composition model $\mathcal{H}$ is implemented with TF-ICON~\cite{lu2023tf}, where the coefficients are set to $\sigma_{ia}=0.2$ for the interaction area and $\sigma_o=0.15$ for the subject mask. The video-to-video translation model $\mathcal{R}$ is realized by AnyV2V~\cite{ku2024anyv2v}, with 50 denoising steps for temporal refinement. 

\noindent\textbf{Common Settings.}
All models are executed on one NVIDIA RTX 6000 Ada GPU with mixed precision, using \texttt{bfloat16} or \texttt{fp16} depending on the model. We standardize all outputs to $512 \times 512$ resolution, 16 frames, and 8 FPS across all experiments. To ensure fairness, each paradigm is evaluated under the same dataset, metrics, and output protocol, while each pretrained model is used with its official default hyperparameters when applicable. We generate 100 videos for each paradigm.

\subsection{Result Analysis}
\begin{figure*}[t]
    \centering
 \includegraphics[width=1\linewidth]{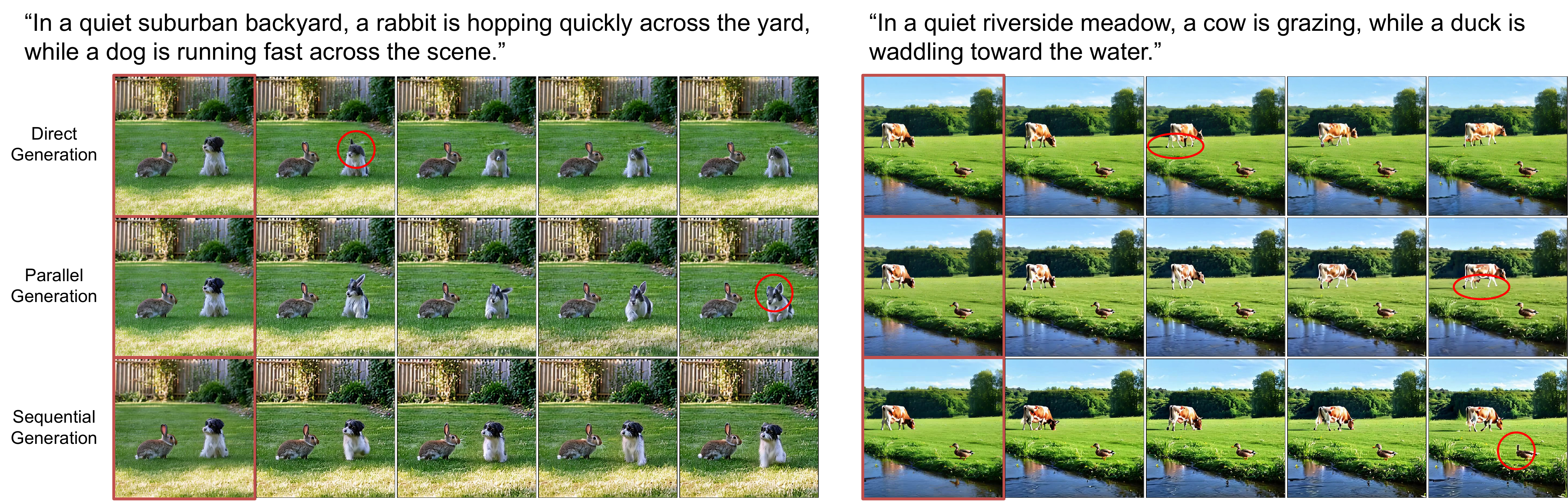}
    \caption{
    Qualitative comparison of the three paradigms for multi-subject I2V generation.
    All paradigms use the same pretrained I2VGen-XL~\cite{zhang2023i2vgen} backbone.
    Columns in each block display the 1st, 3rd, 7th, 11th, and 15th frames of the generated videos.
    The input image $I$ is highlighted with a red box, and the text prompt $T$ is shown above each block.
    Representative unnatural details are highlighted with red circles.
    }
    \Description{Figure described in the caption.}
    \label{fig:comparison}
\end{figure*}

\noindent\textbf{Quantitative comparison.}
Table~\ref{tab:compare_three} compares the three paradigms using I2VGen-XL~\cite{zhang2023i2vgen} and DynamiCrafter~\cite{xing2024dynamicrafter} under matched classifier-free guidance scales and sampling steps.
Sequential generation achieves the highest subject consistency (SC) and background consistency (BC) in all matched configurations, whereas direct generation consistently provides the highest motion smoothness (MS).
This reveals a stable trade-off: progressive insertion preserves accumulated subject and scene context, while single-pass generation avoids temporal disturbances introduced by composition and refinement.

Aesthetic quality (AQ) and imaging quality (IQ) are more dependent on the I2V backbone.
Sequential generation performs the best with DynamiCrafter in most settings, while direct generation generally retains stronger frame-level quality with I2VGen-XL, particularly at 50 sampling steps.
Parallel generation reduces the complexity of each I2V call through independent component generation, but the lack of shared temporal context and the subsequent composition stage prevent this modularity from consistently improving the final metrics.
Overall, no paradigm dominates all dimensions: direct generation favors temporal smoothness, sequential generation favors compositional consistency, and frame-level quality depends on compatibility with the underlying backbone.

\noindent\textbf{Paradigm-level observations.}
Direct generation synthesizes all subjects and their motions jointly through a single pretrained I2V sampling process.
This design consistently produces the highest motion smoothness because it does not require independent video composition or progressive frame-level modification.
However, jointly handling multiple subjects places the full compositional burden on the pretrained backbone, and its SC and BC remain consistently below those of sequential generation.

Parallel generation independently animates the decomposed background and subject components before merging them through video object composition.
Although this design simplifies each individual generation call and permits the component streams to be generated independently, the resulting videos do not share a unified temporal and contextual representation.
Consequently, the reduced per-call synthesis complexity does not consistently translate into stronger final-video metrics, while the subsequent composition process may introduce additional inconsistency.

Sequential generation progressively inserts subjects into an evolving video using frame-level composition and temporal refinement.
By retaining the video context accumulated after each insertion, it consistently achieves the strongest subject and background consistency across both backbones and all evaluated settings.
However, repeated insertion and refinement reduce motion smoothness relative to direct generation.
Their effect on frame-level quality is backbone-dependent: sequential generation is particularly effective with DynamiCrafter, whereas direct generation retains stronger AQ and IQ with I2VGen-XL.
These findings suggest that progressive insertion reliably shifts generation toward stronger compositional consistency, but its overall perceptual benefit depends on its compatibility with the underlying I2V backbone.

\noindent\textbf{Qualitative comparison.}
Figure~\ref{fig:comparison} compares direct, parallel, and sequential generation across two input image--prompt pairs, using the same input conditions and I2VGen-XL~\cite{zhang2023i2vgen} backbone within each example.
In the rabbit--dog example, Direct generation preserves the rabbit reasonably well, but the dog loses its recognizable facial structure and gradually appears as an ambiguous long-eared animal.
Parallel generation also maintains the rabbit, whereas the dog progressively acquires rabbit-like characteristics, indicating cross-subject identity confusion.
Sequential generation preserves the identities of both animals more consistently throughout the sampled frames.
In the cow--duck example, however, all three paradigms exhibit different visual artifacts.
Direct generation produces an anatomically implausible number of cow legs from the early frames, while parallel generation develops a similar leg-count inconsistency toward the end of the video.
Sequential generation maintains the cow and duck identities over most frames, but the duck's head becomes visibly flattened when it turns to a side view in the final frame.

\begin{table}[t] \centering \resizebox{0.45\textwidth}{!}{%
\begin{tabular}{ll|ccc} 
\hline \textbf{Backbone} & \textbf{Metric} & \textbf{Direct} & \textbf{Parallel} & \textbf{Sequential} \\ 
\hline \multirow{2}{*}{I2VGen-XL} & Time (s) $\downarrow$ & 29.48 & 240.25 & 441.45 \\ 
& Peak VRAM (GB) $\downarrow$ & 8.98 & 12.93 & 26.42 \\ 
\hline \multirow{2}{*}{DynamiCrafter} & Time (s) $\downarrow$ & 59.00 & 251.97 & 430.14 \\ 
& Peak VRAM (GB) $\downarrow$ & 10.58 & 14.21 & 23.94 \\ 
\hline \end{tabular}} 
\caption{Computational costs of the three paradigms with two pretrained I2V backbones. Inference time and peak VRAM are measured during end-to-end generation with a single input on an NVIDIA RTX 6000 Ada GPU. Peak VRAM denotes the maximum GPU memory usage observed throughout the complete generation pipeline.}
\label{tab:computational_cost} 
\end{table}

\noindent\textbf{Computational trade-off.}
Table~\ref{tab:computational_cost} reports the end-to-end inference time and peak VRAM consumption of the three paradigms for a single input.
Direct generation achieves the lowest runtime and memory consumption because it requires only one I2V sampling pass using the original multi-subject image and text prompt.
Parallel generation performs separate I2V generation for the background and individual subjects, followed by an additional video object composition stage.
The complete pipeline therefore remains considerably more expensive than direct generation because it requires multiple I2V calls, stores multiple generated streams, and performs an additional composition process.
Sequential generation incurs the highest computational cost because subjects are inserted progressively.
Each insertion requires a DDIM-inversion-based composition operation~\cite{song2020denoising,tumanyan2023plug}, followed by temporal refinement of the updated video.
Consequently, inserting $N$ subjects requires repeated composition and refinement rather than a single post-hoc composition step.
Its higher peak VRAM further reflects the additional video latents and intermediate states maintained during the multi-stage composition and refinement pipeline.
These results reveal a clear runtime--memory--consistency trade-off: direct generation minimizes computational resource usage, whereas sequential generation consumes substantially more time and GPU memory to preserve the context accumulated after each subject insertion.

\begin{figure}[t]
    \centering
    \includegraphics[width=0.95\linewidth]{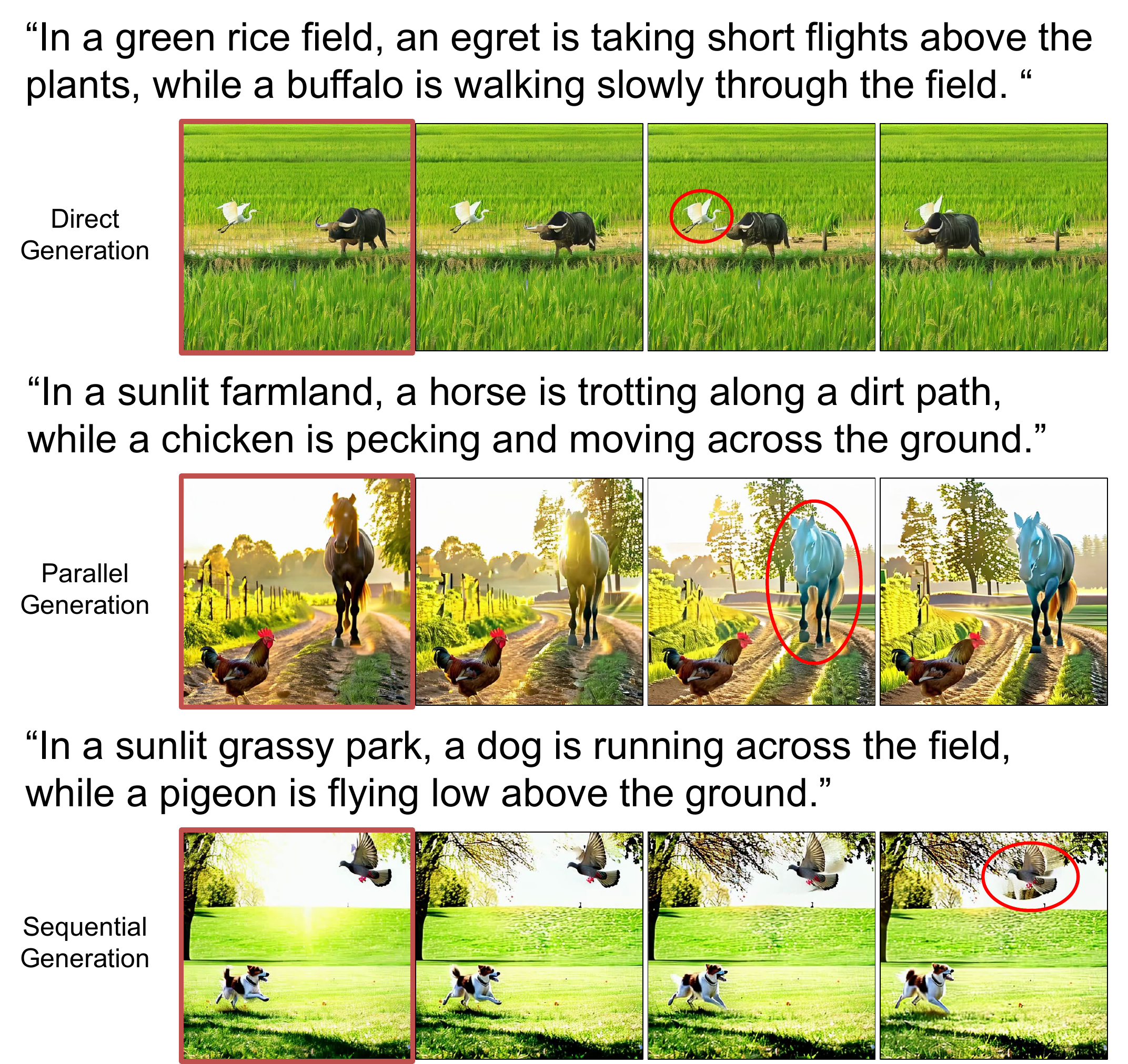}
    \caption{
    Failure cases of the three paradigms for multi-subject I2V generation.
    Columns in each block display the 1st, 3rd, 8th, and 13th frames of the generated videos.
    Representative failure regions are highlighted with red circles.
    }
    \Description{Figure described in the caption.}
    \label{fig:failure_cases}
\end{figure}

\noindent\textbf{Failure case analysis.}
The three paradigms exhibit distinct failure modes reflecting their generation strategies, as illustrated in Fig.~\ref{fig:failure_cases}.
For direct generation, jointly synthesizing multiple subjects with distinct motions may exceed the capability of the pretrained I2V backbone. In the first row, the buffalo moves visibly while the bird remains nearly static, suggesting difficulty in assigning plausible motions to all subjects simultaneously.

Parallel generation inherits errors from independently generated subject-level videos. Errors in appearance or motion are directly propagated to the final composition, as shown by the horse in the second row gradually developing an abnormal blue appearance. Since the composition stage mainly integrates existing streams, it has limited ability to correct such errors.

Sequential generation is sensitive to errors introduced by intermediate processing and insertion. In the third row, inaccurate masks produce residual wing artifacts and an abnormal wing structure, while trajectory or motion estimation errors cause the dog to move backward. Such intermediate errors may persist or accumulate across subsequent stages.

Overall, direct generation is limited by the backbone's generation capability, parallel generation by inherited subject-wise errors, and sequential generation by errors propagated through intermediate modules.

\section{Conclusion and Discussion}
We present a systematic comparison of three paradigms for multi-subject image-to-video generation: direct, parallel, and sequential generation.
Under matched backbones and sampling settings, the results reveal a consistent division of strengths.
Sequential generation achieves the strongest subject and background consistency, while direct generation consistently produces the smoothest motion and requires substantially less inference time.
Frame-level aesthetic and imaging quality are more backbone-dependent, favoring sequential generation with DynamiCrafter but direct generation with I2VGen-XL.
Parallel generation provides a modular decomposition of multi-subject synthesis, although independent generation and post-hoc composition do not consistently improve the final quantitative results.
These indicate that no paradigm is universally optimal: direct generation favors efficiency and temporal smoothness, parallel generation favors modularity, and sequential generation favors compositional consistency at a higher computational cost.

\noindent\textbf{Limitations and Future Work.}
Our study is limited to a relatively small synthetic benchmark, two pretrained I2V backbones, and five automatic evaluation metrics.
The observed trade-offs may therefore change with larger real-world datasets, stronger video generation models, or human evaluation.
Moreover, all paradigms remain constrained by the capability of their pretrained I2V backbones, while parallel and sequential generation additionally depend on the reliability of decomposition and composition modules.
Future work may extend this comparison to more backbones and subjects, introduce human-aligned evaluations, and investigate hybrid strategies that combine the efficiency of direct generation with the context preservation of progressive composition.

\begin{acks}
Research reported in this publication was supported in part by computing resources from the iTiger GPU cluster~\cite{sharif2025cultivatingmultidisciplinaryresearcheducation}, funded by NSF award CNS-2318210 and partially by generous contributions from the College of Arts and Sciences and Information Technology Services at the University of Memphis.
\end{acks}

\bibliographystyle{ACM-Reference-Format}
\bibliography{main}

@inproceedings{ni2024ti2v,
  title={Ti2v-zero: Zero-shot image conditioning for text-to-video diffusion models},
  author={Ni, Haomiao and Egger, Bernhard and Lohit, Suhas and Cherian, Anoop and Wang, Ye and Koike-Akino, Toshiaki and Huang, Sharon X and Marks, Tim K},
  booktitle={Proceedings of the IEEE/CVF Conference on Computer Vision and Pattern Recognition},
  pages={9015--9025},
  year={2024}
}

@inproceedings{ni2023conditional,
  title={Conditional image-to-video generation with latent flow diffusion models},
  author={Ni, Haomiao and Shi, Changhao and Li, Kai and Huang, Sharon X and Min, Martin Renqiang},
  booktitle={Proceedings of the IEEE/CVF Conference on Computer Vision and Pattern Recognition},
  pages={18444--18455},
  year={2023}
}

@inproceedings{hu2022make,
  title={Make it move: controllable image-to-video generation with text descriptions},
  author={Hu, Yaosi and Luo, Chong and Chen, Zhenzhong},
  booktitle={Proceedings of the IEEE/CVF Conference on Computer Vision and Pattern Recognition},
  pages={18219--18228},
  year={2022}
}

@article{kong2024hunyuanvideo,
  title={Hunyuanvideo: A systematic framework for large video generative models},
  author={Kong, Weijie and Tian, Qi and Zhang, Zijian and Min, Rox and Dai, Zuozhuo and Zhou, Jin and Xiong, Jiangfeng and Li, Xin and Wu, Bo and Zhang, Jianwei and others},
  journal={arXiv preprint arXiv:2412.03603},
  year={2024}
}

@article{zhang2023i2vgen,
  title={I2vgen-xl: High-quality image-to-video synthesis via cascaded diffusion models},
  author={Zhang, Shiwei and Wang, Jiayu and Zhang, Yingya and Zhao, Kang and Yuan, Hangjie and Qin, Zhiwu and Wang, Xiang and Zhao, Deli and Zhou, Jingren},
  journal={arXiv preprint arXiv:2311.04145},
  year={2023}
}

@inproceedings{xing2024dynamicrafter,
  title={Dynamicrafter: Animating open-domain images with video diffusion priors},
  author={Xing, Jinbo and Xia, Menghan and Zhang, Yong and Chen, Haoxin and Yu, Wangbo and Liu, Hanyuan and Liu, Gongye and Wang, Xintao and Shan, Ying and Wong, Tien-Tsin},
  booktitle={European Conference on Computer Vision},
  pages={399--417},
  year={2024},
  organization={Springer}
}

@article{yang2024cogvideox,
  title={Cogvideox: Text-to-video diffusion models with an expert transformer},
  author={Yang, Zhuoyi and Teng, Jiayan and Zheng, Wendi and Ding, Ming and Huang, Shiyu and Xu, Jiazheng and Yang, Yuanming and Hong, Wenyi and Zhang, Xiaohan and Feng, Guanyu and others},
  journal={arXiv preprint arXiv:2408.06072},
  year={2024}
}

@article{babaeizadeh2017stochastic,
  title={Stochastic variational video prediction},
  author={Babaeizadeh, Mohammad and Finn, Chelsea and Erhan, Dumitru and Campbell, Roy H and Levine, Sergey},
  journal={arXiv preprint arXiv:1710.11252},
  year={2017}
}

@article{chen2025multi,
  title={Multi-subject Open-set Personalization in Video Generation},
  author={Chen, Tsai-Shien and Siarohin, Aliaksandr and Menapace, Willi and Fang, Yuwei and Lee, Kwot Sin and Skorokhodov, Ivan and Aberman, Kfir and Zhu, Jun-Yan and Yang, Ming-Hsuan and Tulyakov, Sergey},
  journal={arXiv preprint arXiv:2501.06187},
  year={2025}
}

@article{chen2023videodreamer,
  title={Videodreamer: Customized multi-subject text-to-video generation with disen-mix finetuning},
  author={Chen, Hong and Wang, Xin and Zeng, Guanning and Zhang, Yipeng and Zhou, Yuwei and Han, Feilin and Zhu, Wenwu},
  journal={arXiv preprint arXiv:2311.00990},
  year={2023}
}

@article{deng2025cinema,
  title={Cinema: Coherent multi-subject video generation via mllm-based guidance},
  author={Deng, Yufan and Guo, Xun and Wang, Yizhi and Fang, Jacob Zhiyuan and Wang, Angtian and Yuan, Shenghai and Yang, Yiding and Liu, Bo and Huang, Haibin and Ma, Chongyang},
  journal={arXiv preprint arXiv:2503.10391},
  year={2025}
}

@article{wang2024customvideo,
  title={Customvideo: Customizing text-to-video generation with multiple subjects},
  author={Wang, Zhao and Li, Aoxue and Zhu, Lingting and Guo, Yong and Dou, Qi and Li, Zhenguo},
  journal={arXiv preprint arXiv:2401.09962},
  year={2024}
}

@inproceedings{chen2024disenstudio,
  title={Disenstudio: Customized multi-subject text-to-video generation with disentangled spatial control},
  author={Chen, Hong and Wang, Xin and Zhang, Yipeng and Zhou, Yuwei and Zhang, Zeyang and Tang, Siao and Zhu, Wenwu},
  booktitle={Proceedings of the 32nd ACM International Conference on Multimedia},
  pages={3637--3646},
  year={2024}
}

@inproceedings{rombach2022high,
  title={High-resolution image synthesis with latent diffusion models},
  author={Rombach, Robin and Blattmann, Andreas and Lorenz, Dominik and Esser, Patrick and Ommer, Bj{\"o}rn},
  booktitle={Proceedings of the IEEE/CVF Conference on Computer Vision and Pattern Recognition},
  pages={10684--10695},
  year={2022}
}

@article{hacohen2024ltx,
  title={Ltx-video: Realtime video latent diffusion},
  author={HaCohen, Yoav and Chiprut, Nisan and Brazowski, Benny and Shalem, Daniel and Moshe, Dudu and Richardson, Eitan and Levin, Eran and Shiran, Guy and Zabari, Nir and Gordon, Ori and others},
  journal={arXiv preprint arXiv:2501.00103},
  year={2024}
}

@article{zhao2025dreaminsert,
  title={DreamInsert: Zero-Shot Image-to-Video Object Insertion from A Single Image},
  author={Zhao, Qi and Ma, Zhan and Zhou, Pan},
  journal={arXiv preprint arXiv:2503.10342},
  year={2025}
}

@inproceedings{huang2024vbench,
  title={Vbench: Comprehensive benchmark suite for video generative models},
  author={Huang, Ziqi and He, Yinan and Yu, Jiashuo and Zhang, Fan and Si, Chenyang and Jiang, Yuming and Zhang, Yuanhan and Wu, Tianxing and Jin, Qingyang and Chanpaisit, Nattapol and others},
  booktitle={Proceedings of the IEEE/CVF Conference on Computer Vision and Pattern Recognition},
  pages={21807--21818},
  year={2024}
}

@article{ho2022imagen,
  title={Imagen video: High definition video generation with diffusion models},
  author={Ho, Jonathan and Chan, William and Saharia, Chitwan and Whang, Jay and Gao, Ruiqi and Gritsenko, Alexey and Kingma, Diederik P and Poole, Ben and Norouzi, Mohammad and Fleet, David J and others},
  journal={arXiv preprint arXiv:2210.02303},
  year={2022}
}

@inproceedings{liu2024grounding,
  title={Grounding dino: Marrying dino with grounded pre-training for open-set object detection},
  author={Liu, Shilong and Zeng, Zhaoyang and Ren, Tianhe and Li, Feng and Zhang, Hao and Yang, Jie and Jiang, Qing and Li, Chunyuan and Yang, Jianwei and Su, Hang and others},
  booktitle={European Conference on Computer Vision},
  pages={38--55},
  year={2024},
  organization={Springer}
}

@inproceedings{kirillov2023segment,
  title={Segment anything},
  author={Kirillov, Alexander and Mintun, Eric and Ravi, Nikhila and Mao, Hanzi and Rolland, Chloe and Gustafson, Laura and Xiao, Tete and Whitehead, Spencer and Berg, Alexander C and Lo, Wan-Yen and others},
  booktitle={Proceedings of the IEEE/CVF International Conference on Computer Vision},
  pages={4015--4026},
  year={2023}
}

@article{achiam2023gpt,
  title={Gpt-4 technical report},
  author={Achiam, Josh and Adler, Steven and Agarwal, Sandhini and Ahmad, Lama and Akkaya, Ilge and Aleman, Florencia Leoni and Almeida, Diogo and Altenschmidt, Janko and Altman, Sam and Anadkat, Shyamal and others},
  journal={arXiv preprint arXiv:2303.08774},
  year={2023}
}

@inproceedings{lu2023tf,
  title={Tf-icon: Diffusion-based training-free cross-domain image composition},
  author={Lu, Shilin and Liu, Yanzhu and Kong, Adams Wai-Kin},
  booktitle={Proceedings of the IEEE/CVF International Conference on Computer Vision},
  pages={2294--2305},
  year={2023}
}

@article{ku2024anyv2v,
  title={Anyv2v: A tuning-free framework for any video-to-video editing tasks},
  author={Ku, Max and Wei, Cong and Ren, Weiming and Yang, Harry and Chen, Wenhu},
  journal={arXiv preprint arXiv:2403.14468},
  year={2024}
}

@article{wang2024mvoc,
  title={MVOC: a training-free multiple video object composition method with diffusion models},
  author={Wang, Wei and Chen, Yaosen and Liu, Yuegen and Yuan, Qi and Yang, Shubin and Zhang, Yanru},
  journal={arXiv preprint arXiv:2406.15829},
  year={2024}
}

@article{song2020denoising,
  title={Denoising diffusion implicit models},
  author={Song, Jiaming and Meng, Chenlin and Ermon, Stefano},
  journal={arXiv preprint arXiv:2010.02502},
  year={2020}
}

@article{ho2020denoising,
  title={Denoising diffusion probabilistic models},
  author={Ho, Jonathan and Jain, Ajay and Abbeel, Pieter},
  journal={Advances in Neural Information Processing Systems},
  volume={33},
  pages={6840--6851},
  year={2020}
}

@article{song2020score,
  title={Score-based generative modeling through stochastic differential equations},
  author={Song, Yang and Sohl-Dickstein, Jascha and Kingma, Diederik P and Kumar, Abhishek and Ermon, Stefano and Poole, Ben},
  journal={arXiv preprint arXiv:2011.13456},
  year={2020}
}

@inproceedings{tumanyan2023plug,
  title={Plug-and-play diffusion features for text-driven image-to-image translation},
  author={Tumanyan, Narek and Geyer, Michal and Bagon, Shai and Dekel, Tali},
  booktitle={Proceedings of the IEEE/CVF Conference on Computer Vision and Pattern Recognition},
  pages={1921--1930},
  year={2023}
}

@inproceedings{radford2021learning,
  title={Learning transferable visual models from natural language supervision},
  author={Radford, Alec and Kim, Jong Wook and Hallacy, Chris and Ramesh, Aditya and Goh, Gabriel and Agarwal, Sandhini and Sastry, Girish and Askell, Amanda and Mishkin, Pamela and Clark, Jack and others},
  booktitle={International Conference on Machine Learning},
  pages={8748--8763},
  year={2021},
  organization={PMLR}
}

@inproceedings{ravi2025sam,
  title={Sam 2: Segment anything in images and videos},
  author={Ravi, Nikhila and Gabeur, Valentin and Hu, Yuan-Ting and Hu, Ronghang and Ryali, Chaitanya and Ma, Tengyu and Khedr, Haitham and R{\"a}dle, Roman and Rolland, Chloe and Gustafson, Laura and others},
  booktitle={International Conference on Learning Representations},
  volume={2025},
  pages={28085--28128},
  year={2025}
}

@misc{ren2024grounded,
      title={Grounded SAM: Assembling Open-World Models for Diverse Visual Tasks}, 
      author={Tianhe Ren and Shilong Liu and Ailing Zeng and Jing Lin and Kunchang Li and He Cao and Jiayu Chen and Xinyu Huang and Yukang Chen and Feng Yan and Zhaoyang Zeng and Hao Zhang and Feng Li and Jie Yang and Hongyang Li and Qing Jiang and Lei Zhang},
      year={2024},
      eprint={2401.14159},
      archivePrefix={arXiv},
      primaryClass={cs.CV}
}

@misc{sharif2025cultivatingmultidisciplinaryresearcheducation,
      title={Cultivating Multidisciplinary Research and Education on GPU Infrastructure for Mid-South Institutions at the University of Memphis: Practice and Challenge}, 
      author={Mayira Sharif and Guangzeng Han and Weisi Liu and Xiaolei Huang},
      year={2025},
      eprint={2504.14786},
      archivePrefix={arXiv},
      primaryClass={cs.DC},
      url={https://arxiv.org/abs/2504.14786}, 
}

\appendix

\end{document}